%% file: main.tex
\documentclass[runningheads]{llncs}

\usepackage{eccv}

\usepackage{eccvabbrv}

\usepackage{graphicx}
\usepackage{booktabs}
\usepackage{multirow}
\usepackage{xcolor}

\definecolor{cyan}{RGB}{102,184,178}
\definecolor{lightblue}{RGB}{142,181,224}

\usepackage[accsupp]{axessibility}  

\usepackage[breaklinks,colorlinks,citecolor=eccvblue]{hyperref}
\usepackage{hyperref}

\usepackage{orcidlink}
\usepackage{marvosym}

\begin{document}

\title{Towards Open Domain Text-Driven Synthesis of Multi-Person Motions}


\author{Mengyi Shan\inst{1} \orcidlink{0000-0002-1520-5979}
\and Lu Dong \inst{2} \orcidlink{0009-0007-4036-7690}
\and Yutao Han \inst{3} \orcidlink{0000-0002-3867-6925}
\and Yuan Yao \inst{4} \orcidlink{0000-0002-5789-3554}
\and Tao Liu \inst{3}
\and Ifeoma Nwogu \inst{2} \orcidlink{0000-0003-1414-6433}
\and Guo-Jun Qi \inst{5} \orcidlink{0000-0003-3508-1851}
\and Mitch Hill $^{3,{\textrm{\Letter}}}$
} 

\authorrunning{Shan et al.}

\institute{University of Washington, Seattle WA 98105, USA 
\and
University at Buffalo, Buffalo NY 14261, USA
\and
Innopeak Technology, Seattle WA 98004, USA
\and
University of Rochester, Rochester NY 14627, USA
\and
Westlake University, China
\\}

\renewcommand{\thefootnote}{\fnsymbol{footnote}}
\footnotetext{Work done during Mengyi, Lu and Yuan's internship at Innopeak Technology.}
\footnotetext{\textrm{\Letter} Corresponding author and project lead.}

\maketitle

\begin{abstract}
This work aims to generate natural and diverse group motions of multiple humans from textual descriptions. While single-person text-to-motion generation is extensively studied, it remains challenging to synthesize motions for more than one or two subjects from in-the-wild prompts, mainly due to the lack of available datasets. In this work, we curate human pose and motion datasets by estimating pose information from large-scale image and video datasets. Our models use a transformer-based diffusion framework that accommodates multiple datasets with any number of subjects or frames. Experiments explore both generation of multi-person static poses and generation of multi-person motion sequences. To our knowledge, our method is the first to generate multi-subject motion sequences with high diversity and fidelity from a large variety of textual prompts.
\end{abstract}

\input{sections/introduction}
\input{sections/related_works}
\input{sections/dataset}
\input{sections/model}
\input{sections/experiments}
\input{sections/conclusion}

%
%
\bibliographystyle{splncs04}
\bibliography{main}
\end{document}

%% file: sections/introduction.tex
\section{Introduction}
\label{sec:intro}

This work presents a framework for tackling the challenge of generating multi-person motions from a natural language text prompt. Human motion modeling is a widely studied topic with applications spanning areas such as robotics, games, and VR/AR. Conventional approaches for creating computational models of human motion require artists to animate 3D assets~\cite{li20214dcomplete} or an elaborate motion capture process~\cite{moeslund2001mocap}. 
Recently, human motion synthesis with generative models has seen significant progress. Text-driven motion generation~\cite{tevet2023human, guo2022t2m, petrovich22temos, jiang2023motiongpt, tevet2022motionclip, liang2023intergen, tanaka2023interaction}, greatly increases the efficiency, flexibility and accessibility of motion animation. Nonetheless, prior works are limited to single-person or two-person motions, and often are not compatible with prompts that extend beyond a restricted distribution. These limitations are primarily due to available motion data. Prior works are confined to training models with single-person~\cite{guo2022t2m} or two-person~\cite{liang2023intergen} motion datasets with moderately diverse prompt distributions.

\begin{figure}
    \centering
    \includegraphics[width=\textwidth]{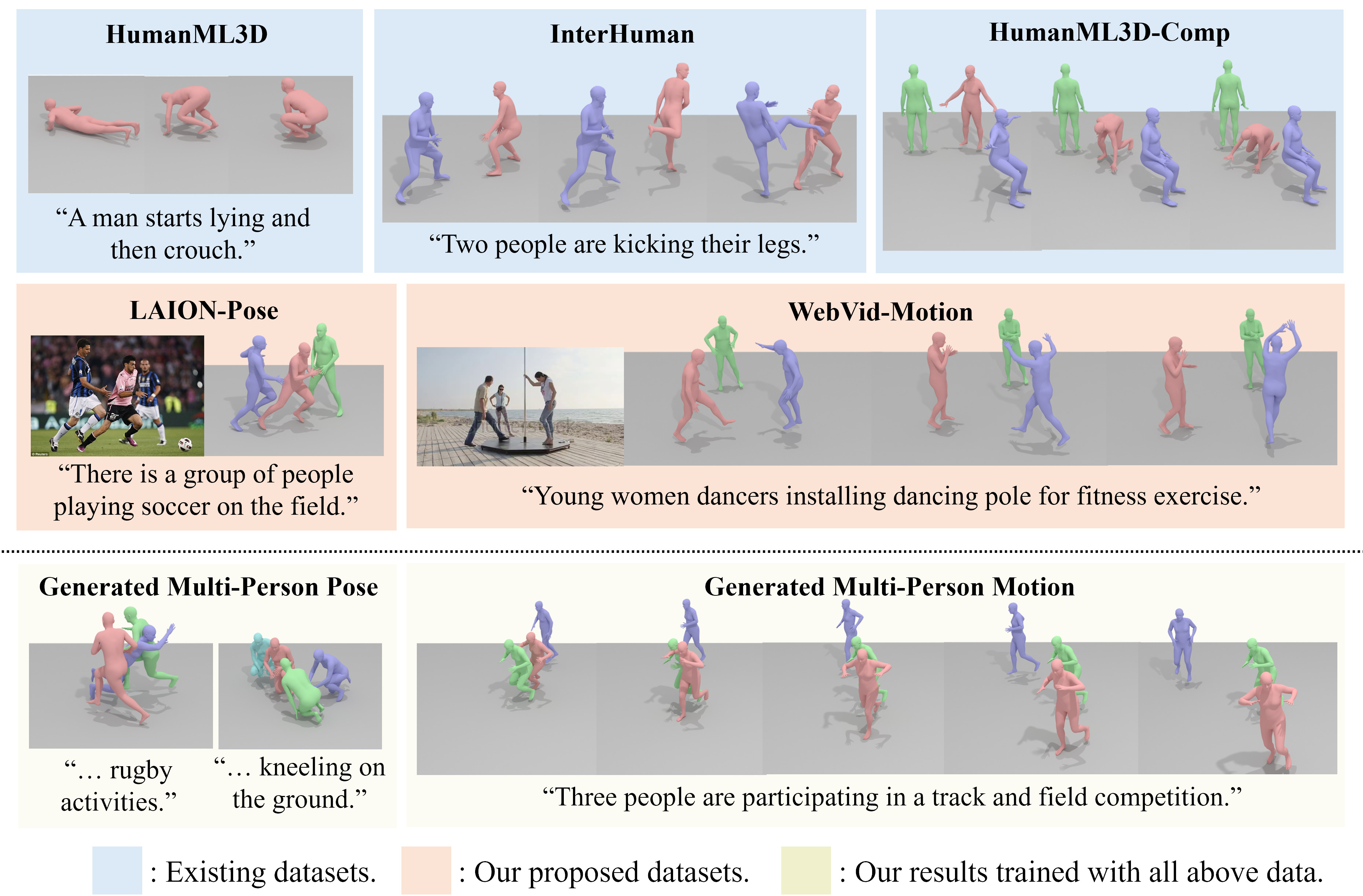}
    \caption{We jointly train with multiple data sources including motion capture data and pose/motion extracted from image/video datasets. The model generates motion sequences from text for an arbitrary number of subjects. }
    \label{fig:teaser}
\end{figure}

We address the challenge of multi-person motion modeling by introducing novel datasets which provide multi-person poses and motions along with text descriptions. Given the difficulties of multi-person motion capture, our strategy is instead to build upon recent advances for estimating pose and motions from image and video sources. In particular, we leverage human pose estimation methods BEV~\cite{sun2022bev, sun2023trace} and TRACE~\cite{sun2023trace} to extract multi-person static pose and dynamic motion from large-scale image dataset LAION-400M~\cite{schuhmann2021laion} and video dataset WebVid-10M~\cite{bain21webvid}. This results in our LAION-Pose dataset with 8 million \texttt{(image, pose, text)} tuples and our WebVid-Motion dataset with 3,500 \texttt{(video, motion, text)} tuples. Sample text descriptions span a diverse variety of poses and motions that occur in open-domain images and videos.

To accommodate multiple data sources (single/multi subject, single/multi frame), we represent all samples in a common format given by the SMPL~\cite{loper2023smpl} parameters of each frame. Inspired by video generation architectures~\cite{ho2022video}, our multi-person motion network uses interleaved pose and motion transformer encoder layers~\cite{vaswani2017attention}. Before each pose/motion layer, the temporal/subject dimension of the sample is reshaped into the batch dimension. After each layer, the sample is reshaped to its original format. Consequently, the pose and motion layers focus on generating plausible group poses for each frame, and temporal connection among frames for each subject, respectively.

We adopt a denoising diffusion framework for motion modeling~\cite{ho2020denoising, tevet2023human}. Our proposed method uses a two-stage pipeline. The first stage model produces a single frame containing the poses of multiple people. This sample is used as a condition for the second stage model, which will generate a motion sequence that has the first-stage pose sample as the middle frame. Both stages use the same text prompt for generation. This achieves the effect of generating a pose and animating it over time. During sampling time, we also use pose and single-person motion models as guidance terms to further boost the quality of the multi-person motion results.

Models are evaluated in a decomposed manner by measuring the quality of each multi-person frame and each single-person motion sequence. We train feature extractors for pose and motion in the SMPL format with LAION-Pose and HumanML3D~\cite{guo2022t2m} data following CLIP training~\cite{radford2021learning}. 




Our main contributions can be summarized as follows:
\begin{itemize}
    \item We present the first model to generate multi-person motion sequences given open-domain textual prompts with an arbitrary number of subjects.
    \item We introduce LAION-Pose and WebVid-Motion, the first large-scale 3D datasets of text-annotated multi-person poses/motions collected from in-the-wild images and videos. 
    \item We design a factorized method to evaluate our results in the absence of ground-truth multi-person motion data by building contrastive encoder backbones for text and pose/motion, and demonstrate that our method outperforms existing methods qualitatively and quantitatively.
\end{itemize}

%% file: sections/related_works.tex
\section{Related Works}
\label{sec:related_works}

\noindent \textbf{Human Mesh Recovery.}
Human mesh recovery aims to jointly estimate 3D human body poses and shapes from images or videos. Recent advancements in body parametric models, such as SMPL~\cite{loper2023smpl}, facilitate such development in diverse human-centric tasks including 3D human body reconstruction~\cite{hmrKanazawa17}, animation~\cite{yuan2023gavatar}, and pose estimation~\cite{kocabas2021pare, yuan2021simpoe}.
A variety of works explore single-person human body recovery from single images~\cite{kanazawa2018hmr,kolotouros2019GraphCMR,yao2019densebody,kolotouros2019spin,zhang2021pymaf,in2021graphormer,kolotouros2021prohmr,li2022cliff,goel2023hmr20, zanfir2021hund, kocabas2021pare, yu2021skeleton2mesh} and videos~\cite{kanazawa2019hmmr,doersch2019sim2real, sun2019dsd, kocabas2020vibe,choi2021tcmr, yuan2021simpoe,li2022dtsvibe, wei2022mpsnet}, as well as multi-person mesh recovery scenarios~\cite{choi2022crowdnet,fieraru2021remips,khirodkar2022ochmr,zanfir2018mubynet, sun2021romp,sun2022bev,qiu2023psvt,rajasegaran2021hmar, yuan2022glamr,sun2023trace}.
In particular, ROMP~\cite{sun2021romp} regresses multiple 3D meshes along with the horizontal/vertical locations and a rough depth estimate from a single image. BEV~\cite{sun2022bev} improves upon ROMP by enabling more precise multi-person depth estimation. TRACE~\cite{sun2023trace} extends the setup of BEV and supports motion regression with frame-wise consistency from a dynamic camera video. 
Our paper builds upon 3D human body models and estimation tools to recover multi-person mesh poses from large-scale image and video datasets.

\noindent \textbf{Conditional Motion Generation.} 
Generating realistic 3D human motions with conditional signals is an active research area. Reference motions can be used as conditions to predict future motions~\cite{barsoum2018hp,  fragkiadaki2015recurrent, hernandez2019human, mao2020history, martinez2017human}, intermediate motions~\cite{duan2021singleshot, harvey2020inbetween, harvey2021recurrent}, and motion variations~\cite{li2022ganimator}.
Motion generation can also be conditioned by action class~\cite{petrovich2021action}, audio~\cite{ng2022learning, ng2024audio}, music~\cite{zhao2023taming, le2023music,ma2022pretrained}, and natural language, which is the focus of this work.
Text-conditioned motion generation can be accomplished by a joint embedding of language and pose~\cite{tevet2022motionclip}, VAE structure~\cite{petrovich22temos, guo2022t2m, zhai2023language, yang2023omnimotiongpt}, or VQ-VAE~\cite{guo2022tm2t, jiang2023motiongpt,zhang2023t2m}. Several recent works apply diffusion models~\cite{ho2020denoising} for text-to-motion generation~\cite{tevet2023human, zhang2023remodiffuse, chen2023executing, yuan2023physdiff}. We build upon the MDM~\cite{tevet2023human} framework that uses transformer encoders and extend the architecture to accommodate the joint training scenario with more than one subject.
%

\noindent \textbf{Multi-person Motion Generation.} Despite the success of single-person motion models, generating group motions of more than one person remains challenging.
One direction of research explores multi-subject motion prediction based on past 3D skeletons~\cite{wang2021multi, guo2022multi, tanke2023social, maheshwari2022mugl}. 
In the area of text-driven multi-person motion, ComMDM~\cite{shafir2023human} proposed to use 
a generative prior coupled with a streamlined communication block to facilitate coordinated interaction. 
RIG~\cite{tanaka2023role} attempts to recover 3D interaction motions from a noisy depth dataset~\cite{liu2019ntu} and translates the motion labels into sentences. 
InterGen~\cite{liang2023intergen} introduces two collaborative transformer-based denoisers with a mutual attention mechanism.
InterControl~\cite{wang2023intercontrol} enables flexible spatial control over each joint to generate plausible interactions.
These works focus predominantly on interactions involving only two individuals and are often incompatible with open domain text prompts. Our model generates multi-person motion directly from textual descriptions for one, two, and more than two people, while also exhibiting flexible text generalization abilities learned from in-the-wild datasets.

\noindent \textbf{Human Motion Datasets.} 
Datasets play a crucial role in propelling motion generation research forward. 
Action label datasets~\cite{punnakkal2021babel, liu2019ntu}  
and text annotation datasets~\cite{plappert2016kit, guo2022t2m} support the advancement of single-person motion generation~\cite{chen2023executing}. However, models trained with single-motion datasets will have difficulty generalizing to group motions.
Meanwhile, various multi-person motion datasets have been developed including 4DAssociation~\cite{zhang20204d},
UMPM~\cite{van2011umpm}, 
3DPW~\cite{von2018recovering},
MuPoTS-3D~\cite{mehta2018mupots},
ExPI~\cite{guo2022multi},
CMU-Panoptic~\cite{joo2019huggling},
You2Me~\cite{ng2020you2me}. 
Among these datasets, only 3DPW is compatible with the SMPL format which is a common starting point of current motion generation models. ComMDM~\cite{shafir2023human} contributed textual annotations to 3DPW, but the resulting dataset contains only 27 two-person motion sequences. 
InterHuman~\cite{liang2023intergen} is a recently introduced dataset with diverse multi-person motions with textual annotations. However, it remains constrained to interactions between two individuals. To the best of our knowledge, we contribute the first large-scale datasets of text-annotated 3D poses and motions with more than two people.

%% file: sections/dataset.tex
\section{Dataset}
\label{sec:dataset}

\begin{figure}[!ht]
    \centering
    \includegraphics[width=\textwidth]{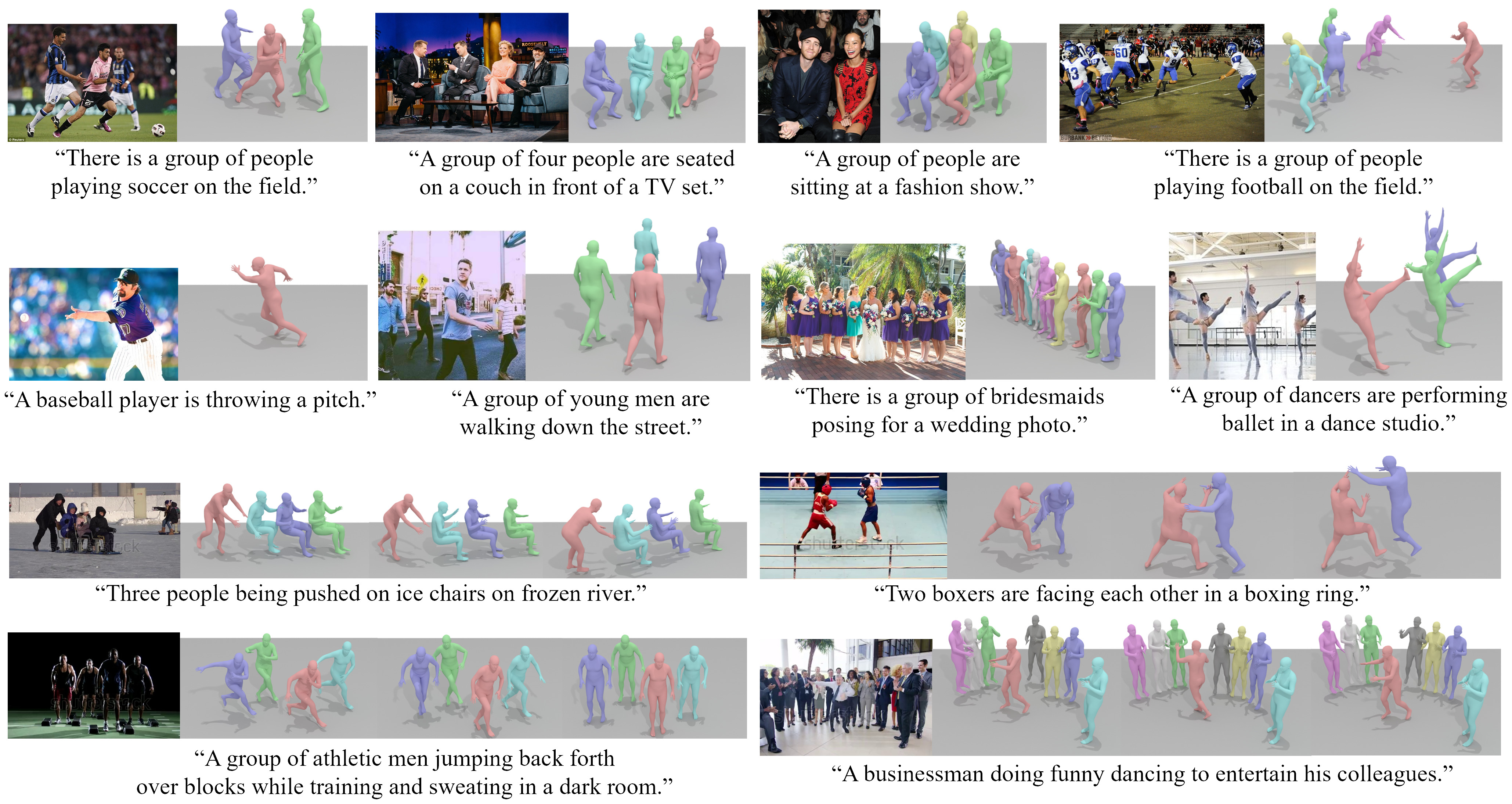}
    \caption{Dataset visualizations. Top 2 rows: LAION-Pose dataset. Left is original image from LAION-400M~\cite{schuhmann2021laion}, right is BEV~\cite{sun2022bev} detection. Bottom 2 rows: Webvid-Motion dataset. Left is original video first frame from WebVid-10M~\cite{bain21webvid}, right is the motion sequence estimated by TRACE~\cite{sun2023trace} visualized from a different camera angle.}
    \label{fig:dataset}
\end{figure}

To address the data scarcity problem in multi-person domain, we introduce two datasets: LAION-Pose and WebVid-Motion, containing \texttt{(image, pose, text)} and \texttt{(video, motion, text)} tuples extracted from in-the-wild images and videos. We present our data processing techniques of LAION-Pose in \cref{sec:laion-pose} and of WebVid-Motion in \cref{sec:webvid-motion}. See \cref{fig:dataset} for selected examples. 

\subsection{LAION-Pose}
\label{sec:laion-pose}

We apply BEV~\cite{sun2022bev} to each image in LAION-400M. BEV estimates human global translation, SMPL joint rotation, and SMPL shape information for an arbitrary number of people in a monocular image. If BEV cannot detect a human occurrence in an image, the image sample is ignored. BEV often provides reasonable mesh estimates for images with clearly visible poses. Nonetheless there are also many circumstances where it does not produce satisfactory results. Subsequent filtering steps are performed to retain predominantly BEV meshes that accurately predict ground truth poses in an image. The filtering stages start with coarse criteria and become progressively more fine. 

\begin{itemize}
\item \textbf{Person Detection.} Detectron-2~\cite{wu2019detectron2} is used to identify whether people are in an image. Samples without any positive detections are removed.

\item \textbf{Mesh Completion.} We retain all meshes such that over 85\% of the mesh falls in the image frame and discard those predictions where human poses are only partially contained within the image frame.


\item \textbf{Mesh De-duplication.}
BEV sometimes predict duplicate, overlaid meshes from the monocular image perspective. Duplicates are detected by projecting the mesh vertices to the image plane and measuring the overlap. One sample in each pair with over 25\% IoU overlap is discarded.

\item \textbf{Hand-Crafted Prompt Filter.}
We hand-craft a set of filtering prompts such as \emph{``a DVD cover"}. We discard all image samples whose CLIP~\cite{radford2021learning} similarity with a filter prompt is higher than a threshold


\item \textbf{Few-Shot Filter.} After the previous filtering steps, we manually annotate 1000 random data samples with binary labels indicating whether the BEV prediction were a visually acceptable match to the reference image. A logistic regression classifier is then trained using the quality annotations and the CLIP image features of the corresponding images.
We set a threshold that such that 90\% of model selections in a validation set are annotated as high-quality to ensure that mostly good samples are kept. 
\end{itemize}

\noindent After data filtering, we are left with around 8 million \texttt{(image, pose, text)} triplets which are the raw form of our dataset. The final dataset is created by refining pose samples and generating more informative text captions.

\begin{itemize}
\item \textbf{Vertical Height Adjustment.}
BEV faces inherently limitations when estimating spatial relationships among multiple objects due to monocular ambiguity. 
In most cases, editing group poses so that all individuals have a consistent vertical height preserves the essence of the pose description while making group pose appearance more natural. 
A smaller dataset without height adjustment is also kept to learn cases where relative heights are meaningful.

\item \textbf{Mesh Separation.}
We correct mesh overlap by optimizing each pair of SMPL parameters to minimize a mesh collision loss. We compute the degree of overlap between each pair as the sum of SDF (signed distance function) from all vertices in the first mesh to the second mesh, and then run gradient descent to back-propagate through the SMPL layers to optimize for a fixed number of 25 steps. The resulting meshes have no or very little overlap.

\item \textbf{InstructBLIP Captioning.}
LAION-400M~\cite{schuhmann2021laion} captions are not informative for  pose and motion modeling because they almost always focus on appearance, style, or factual metadata of the image instead of human actions. We use InstructBLIP~\cite{dai2023instructblip} instead as an image captioning tool. We prompt InstructBLIP with the request \emph{``describe the person or group of people's action and body poses in the image''}. This simple instruction yields robust captions that are almost always more suitable pose and action descriptions than the LAION-400M text captions. 

\end{itemize}




\subsection{WebVid-Motion}
\label{sec:webvid-motion}

\noindent \textbf{Person Detection.}
We first filter WebVid-10M~\cite{schuhmann2021laion} by applying Detectron-2~\cite{wu2019detectron2} on the middle frame of each video, and only keep videos with at least two detected humans. This works in most cases since WebVid-10M videos tend to have limited camera movement and scene changes.

\noindent \textbf{TRACE Estimation.}
Next, we apply TRACE~\cite{sun2023trace} to filtered videos to predict per frame global translation, joint rotation, and shape information for an arbitrary number of people in a single-view video with dynamic camera. We apply a Gaussian filter with scale $s=1$ to smooth out the motion jittering effects.

\noindent \textbf{Motion Grouping.} 
According to the temporal nature of video, there might be multiple subjects appearing and disappearing through time, and there might be multiple useful motion clips in one video sample. To obtain multi-person motion samples from the raw TRACE outputs, we select all clips where at least two people appear in the video simultaneously for at least 30 frames.

\noindent \textbf{Manual Selection.}
The data is manually inspected to select 3,500 samples with the best fidelity and the most interesting group or interactive behavior. 
Manual selection filters out motions with highly unrealistic translations due to the ambiguity of monocular video, and removes duplicated and static motions.

\noindent \textbf{InstructBLIP Captioning.}
Similar to the image case, we use InstructBLIP~\cite{dai2023instructblip} as a video captioning tool by instructing it to \emph{``describe the person or group of people's action and body poses in the image''} for the middle frame of the video.  

%% file: sections/model.tex
\section{Model}
\label{sec:model}

\begin{figure}
    \centering
    \includegraphics[width=\textwidth]{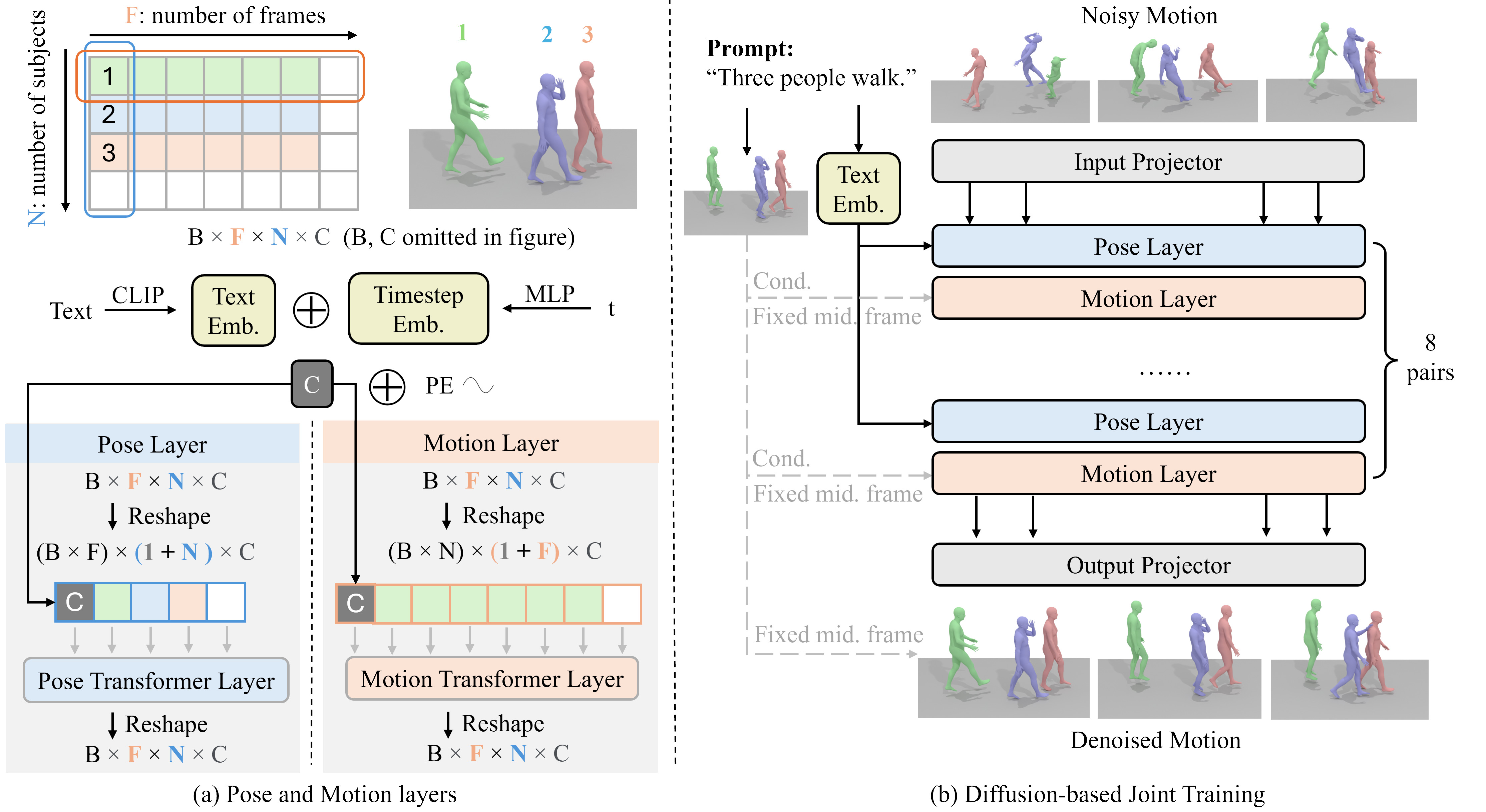}
    \caption{Our model is a diffusion framework consisting of interleaving pose and motion layers. At each pose/motion layer, we reshape the temporal/subject dimension into the batch dimension so that the layer focuses on generating per frame subject interaction and per-subject temporal movements respectively. Each layer is implemented as a transformer encoder. Diffusion time steps and text or pose conditions are encoded and summed up as a condition token concatenated to the beginning of the sequence.}
    \label{fig:model}
\end{figure}

Our goal is to generate high-quality multi-person motion sequences given a textual description. We leverage a transformer architecture made of interleaved pose and motion layers that facilitate plausible frame-wise poses and temporal movements, respectively. \cref{sec:representation} introduces the representation we use for pose and motion. \cref{sec:training} describe the design of our motion and pose layers as well as the two-stage joint training architecture. \cref{sec:sampling} presents our sampling strategy with specific guidance term. The model overview is shown in \cref{fig:model}.

\subsection{Motion Representation}
\label{sec:representation}

We base our motion representation on the widely used SMPL~\cite{loper2023smpl} format, which describe the human body in $24 \times 3$ SMPL $\theta$ parameters for pose and 11 SMPL $\beta$ parameters for motion. For training stability, we follow common practice and turn the axis angle rotation vectors into 6D representation, and add in a global translation $t \in \mathbb{R}^3$ to model movements and spatial relationship among multiple subjects. A single person pose $y \in \mathbb{R}^{158}$ is defined to be the concatenation of $\beta \in \mathbb{R}^{11}$, $\theta \in \mathbb{R}^{24 \times 6}$ and $t \in \mathbb{R}^3$. A multi-person motion is defined as a set of poses $x = {\{y^{n, f}\}_{n=1 f=1}^{N\mkern17mu F}}$ where $F \ge 1$ and $N \ge 1$ are the number of frames and people that comprise the motion. In practice, differences in motion lengths/number of poses are handled by padding to a maximum number of frames/poses and masking attention for pad tokens. Special cases include $F=1$, which corresponds to a multiple poses over a single frame, and $N=1$, which corresponds to single-person motion.

Note that we refrain from using the popular HumanML3D~\cite{guo2022t2m} format for two reasons. First, it contains velocity which is meaningless for pose data and thus not adaptable to our joint training strategy. Second, it canonicalizes joint positions and velocities to the root frame, and thus lose spatial interaction information for multi-person scenarios (as also mentioned in~\cite{liang2023intergen}).

\subsection{Training}
\label{sec:training}

\noindent \textbf{Diffusion Model Objective.}
Denoising diffusion probabilistic~\cite{ho2020denoising} models are a class of generative models that aim to approximate a data distribution through a progressive denoising process. Our forward diffusion procedure can be modeled as a Markov noising process, $\{x_t^{1: N, 1:F}\}^T_{t=0}$ where $t$ is the diffusion time step, $x_0^{1: N, 1:F}$ is drawn from the training data distribution, and 
\[q\left(x_t^{1: N, 1: F} \mid x_{t-1}^{1: N, 1:F}\right)=\mathcal{N}\left(\sqrt{\alpha_t} x_{t-1}^{1: N. 1:F},\left(1-\alpha_t\right) I\right)\] where $\alpha_t \in(0,1)$ are constants. 
Our motion generation model is defined as learning the reverse process of the Markov Chain, gradually cleaning up $x_T$ to get $p\left(x_0 \mid c\right)$ with condition $c$. Following MDM~\cite{tevet2023human}, we design our model to predict the original signal $x_0$ instead of noise $\epsilon_0$ with the simple loss in~\cite{ho2020denoising}:
\[\mathcal{L}=E_{x_0 \sim q\left(x_0 \mid c\right), t \sim[1, T], x_t \sim q_t(x_t | x_0, c)}\left[\left\|x_0-G\left(x_t, t, c\right)\right\|_2^2\right]\]
where $G$ is the motion generation model described in the next section and $q_t$ give the forward process condition distributions. During training we randomly mask out the textual condition by setting the textual embedding to zero for 10\% of samples to enable classifier-free guidance at inference time.

\noindent \textbf{Pose and Motion Layers.} The components of our model follow designs proposed by MDM~\cite{tevet2023human}. Each layer is implemented in a straightforward transformer~\cite{vaswani2017attention} encoder architecture. The transformer layer can used masked attention to remove the influence of padded tokens and thus can handle motions of arbitrary lengths or with arbitrary number of subjects. Pose layers are applied frame-wise, learning relations between poses and locations of multiple subjects for each single frame. Motion layers are applied subject-wise, learning plausible temporal movement of each person. We project the noise time-step $t$ and the CLIP~\cite{radford2021learning} embedding of text to the transformer dimension by separate feed-forward networks, then sum them up to yield the condition token. The token is then appended to the start of each sequence. 


\noindent \textbf{Joint Architecture.} Our joint training architecture (\cref{fig:model} right) uses pairs of pose and motion layers. We start with data padded to the shape $B \times F \times N \times C$ where $B$ is the batch size, $F$ is the maximum number of frames, $N$ is the maximum number of subjects, and $C=158$ is the pose channel. We apply a linear layer that projects the data into a high dimension ($C'=512$). The batch data then goes through each of the pose and motion layers in order. Note that we only apply text condition to pose layers so that single-frame and multi-frame samples have a consistent text conditioning mechanism.

Before entering a pose layer, we reshape the temporal dimension into the batch dimension so that the data shape becomes $(B \times F) \times N \times C'$. We apply pose layer to the subject ($N$) dimension so that it is able to learn the per-frame interaction among multiple subjects. Similarly, before entering a motion layer, we reshape the subject dimension into the batch dimension so that the data shape becomes $(B \times N) \times F \times C'$. We apply motion layer to the temporal ($F$) dimension so that it learns the per-subject movements through time. After each transformer encoder layer, we discard the first output token which corresponds to the conditional signal, and feed the rest into the next layer with corresponding reshaping. After the last layer, we reshape back to $B \times F \times N \times C'$ and project back to the pose dimension $C$ with a linear layer. For simplicity, we omit the additional condition token in the above discussion of reshaping.

Our joint architecture takes both text and a reference pose frame as conditions. The reference pose represents the middle frame of the sequence being generated and the joint model is intended to extend the reference frame forward and backward in time. The pose conditioning is implemented in an analogous way to the text conditioning of pose layers. The reference pose is concatenated with the positional embedding of the middle frame, fed through an MLP, and added to a separate copy of the timestep embedding vector. This embedding is appended as the first token before motion layers.

\noindent \textbf{Two Stage Training.} We train the model in a two-stage manner. First, we train a text-to-pose model with only single-frame multi-person pose data. This model follows the structure of MDM, except for lack of positional encoding. After training, the model can generate single frame pose samples which will be animated over time. The multi-person motion model is initialized by inserting a motion layer after each pose layer and freezing the pose layers. It takes both text and the the middle frame of data motions as conditions during training. Single-frame inputs are assigned a null pose condition, so that the new motion layers can still learn from single-frame samples. Our approach is inspired by similar techniques in the video generation literature such as AnimateDiff~\cite{guo2023animatediff} which insert new temporal layers between frozen spatial layers to preserve certain behaviors of the spatial layers over time. In our case, the we want to preserve the flexible text conditioning learned from our large-scale pose dataset.


\subsection{Sampling}
\label{sec:sampling}

\noindent \textbf{Diffusion Model Sampling.} We sample from our model in an iterative manner following DDPM~\cite{ho2020denoising}. Given a time step $t$ at sampling time, we predict the starting data $\hat{x}_0=G\left(x_t, t, c\right)$ and noise it back to $x_{t-1}$. We repeat this process until we reach $t=0$ from $t=T$. Classifier-free guidance is used to encourage better text alignment by modifying model predictions with a guidance scale $s>1$ according to $G_s\left(x_t, t, c\right)=G\left(x_t, t, \emptyset\right)+s \cdot\left(G\left(x_t, t, c\right)-G\left(x_t, t, \emptyset\right)\right)$.

\noindent \textbf{Two Stage Sampling.} At inference time, we draw samples from our model in two stages. A single-frame multi-person sample is drawn from our pose-only model given a text prompt. The text prompt and the single-frame pose are then used to generate a multi-person motion where the single-frame pose condition condition serves as the center motion frame.

\noindent \textbf{Pose/Motion Guidance.} We additionally leverage separate text-to-pose and text-to-single-person motion models as guidance terms for our multi-person generation. It is relatively straightforward to train single-frame or single-person models with high-quality results, and we may push our multi-person motion generations towards them frame-wise and/or subject-wise. With scale $s_p \ge 0$ and $s_m \ge 0$ for pose and motion models, the guided model can be expressed as 
\[
G_{s_p, s_m}(x_t, t, c )=(1 - s_p - s_m) \cdot G (x_t, t, c) + s_p \cdot G_p (x_t, t, c) +\mathrm{s}_m \cdot G_m (x_t, c)
\]
\noindent where $s_p + s_m \le 1$ and $G_p$, $G_m$ are separate text-to-pose and unconditional single-person motion generators, with frame/subject dimensions put along the batch dimension for pose/motion models respectively. We can easily control the scales to make the results better in motion quality (higher $s_m$), or better in subject interaction (higher $s_p$).

%% file: sections/experiments.tex
\section{Experiments}
\label{sec: experiments}

\subsection{Dataset}
We use five datasets in our joint training: one pose and four motion datasets. All samples are shifted by a global translation such that the average global coordinate of each person in the horizontal plane is the origin for the center motion frame. Samples are augmented by randomly rotating the group motions around the vertical axis passing through the origin.

\noindent \textbf{LAION-Pose} (\cref{sec:laion-pose}) has 8 million pairs of multi-person poses and texts.

\noindent \textbf{WebVid-Motion} (\cref{sec:webvid-motion}) has 2,900 pairs of multi-person motions and texts. 

\noindent \textbf{HumanML3D~\cite{guo2022t2m}} is the most widely used text-motion dataset containing 14,616 single-person motions. We discard the HumanAct12 subset and use it in the original AMASS~\cite{mahmood2019amass} SMPL~\cite{loper2023smpl} format as explained in \cref{sec:representation}.

\noindent \textbf{HumanML3D-Comp} is a synthetic dataset that we compose by arbitrarily selecting 2 to 6 motion sequences from HumanML3D and putting them together in a 3D space with randomly initialized starting translation. Generated samples are checked to ensure there is no geometric collision in meshes between subjects. Since there is no text describing the resulting group motion, they are paired with empty text as unconditional multi-person motion samples.

\noindent \textbf{InterHuman~\cite{liang2023intergen}} is a two-subject interactive motion dataset with a diverse range of around 8,000 text-annotated motions.

\subsection{Implementation Details}
Our model is trained with maximum 61 frames and 10 subjects. The dataset is an amalgamation of the data sources with a split ratio (LP 50\%, WVM 10\%, HML 15\%, HML-C 10\%, IH 15\%). First stage training uses a random data frame while second stage training uses up to 61 frames from a motion sequence. Longer sequences are randomly truncated. Each pose and motion layer is implemented as a transformer encoder layer with latent dimension of 512. The first stage model consists of 8 pose layers and the second stage model consists of 8 pairs of pose and motion layers. The input and output projectors are linear layers. The first stage model is trained on 4 GPUs for 500k steps over 1 day and the second stage model is trained on 8 A100 GPUs for 250K steps over 2 days.  

\subsection{Evaluation Setting}

\noindent \textbf{Number of Poses Determination.} We use language model Mistral~\cite{jiang2023mistral} to decide the number of subjects from a given textual description by asking it \emph{``how many subjects appear in this description''}. 


\noindent \textbf{Decomposed Evaluation.} Existing motion generation models typically evaluate their results with the text-motion shared embedding and metrics proposed by~\cite{guo2022t2m}. See the supplementary material for a description of each metric. In the shortage of ground truth data for training a multi-person motion encoder, we choose to implement a decomposed evaluation mechanism. Specifically, we evaluate our generative metrics on each single frame's multi-person pose result as well as each single subject's single-person motion result. We train two feature encoders with \texttt{(text, pose)} pairs from LAION-Pose and \texttt{(text, motion)} pairs from HumanML3D in the original SMPL format following CLIP~\cite{radford2021learning} training. In practice, to speed up the evaluation we only calculate pose metrics on sparse frames that are separated by 14 frames, yielding 5 evaluation points that are the first, last, and middle quarters of the frame sequence.

We use a withheld validation set of LAION-Pose samples for pose metric and as evaluation prompts covering a wide range of natural descriptions. R-Precision and Similarity are not applicable for motions generated from LAION-Pose prompts because the motion encoder trained with HumanML3D is not compatible with LAION-Pose prompts. Instead, we primarily focus on motion FID calculated using HumanML3D reference motions. This serves as a rough measure of motion realism that describes how well the distribution of generated motions match the distribution of ground truth single-person motions.

\noindent \textbf{Baselines.} To our knowledge, there is no existing work that generates motion with an arbitrary number of subjects to compare with. As naive baselines, we use \textit{Pose-Only} models which lack temporal connections or \textit{Motion-Only} model which lack pose connections. The baselines both provide evidence of importance of joint modeling as well as provide rough upper bounds for the quality metrics that are achievable for frame-wise and person-wise generation in our setup.
In partciular, our \textit{Pose-Only} baseline takes a sample from our first stage model and keeps the pose static throughout the whole sequence. As a \textit{Motion-Only} baseline, we fix the middle frame as the pose result from the first stage, and animate each single subject independently with a model containing only motion layers. We additionally experiment with sampling two subjects motion to compare with two-person motion baselines RIG~\cite{tanaka2023interaction}, InterGen~\cite{liang2023intergen} and ComMDM~\cite{shafir2023human}. 

\subsection{Qualitative Evaluation}

\begin{figure}[ht!]
    \centering
    \includegraphics[width=0.95\textwidth]{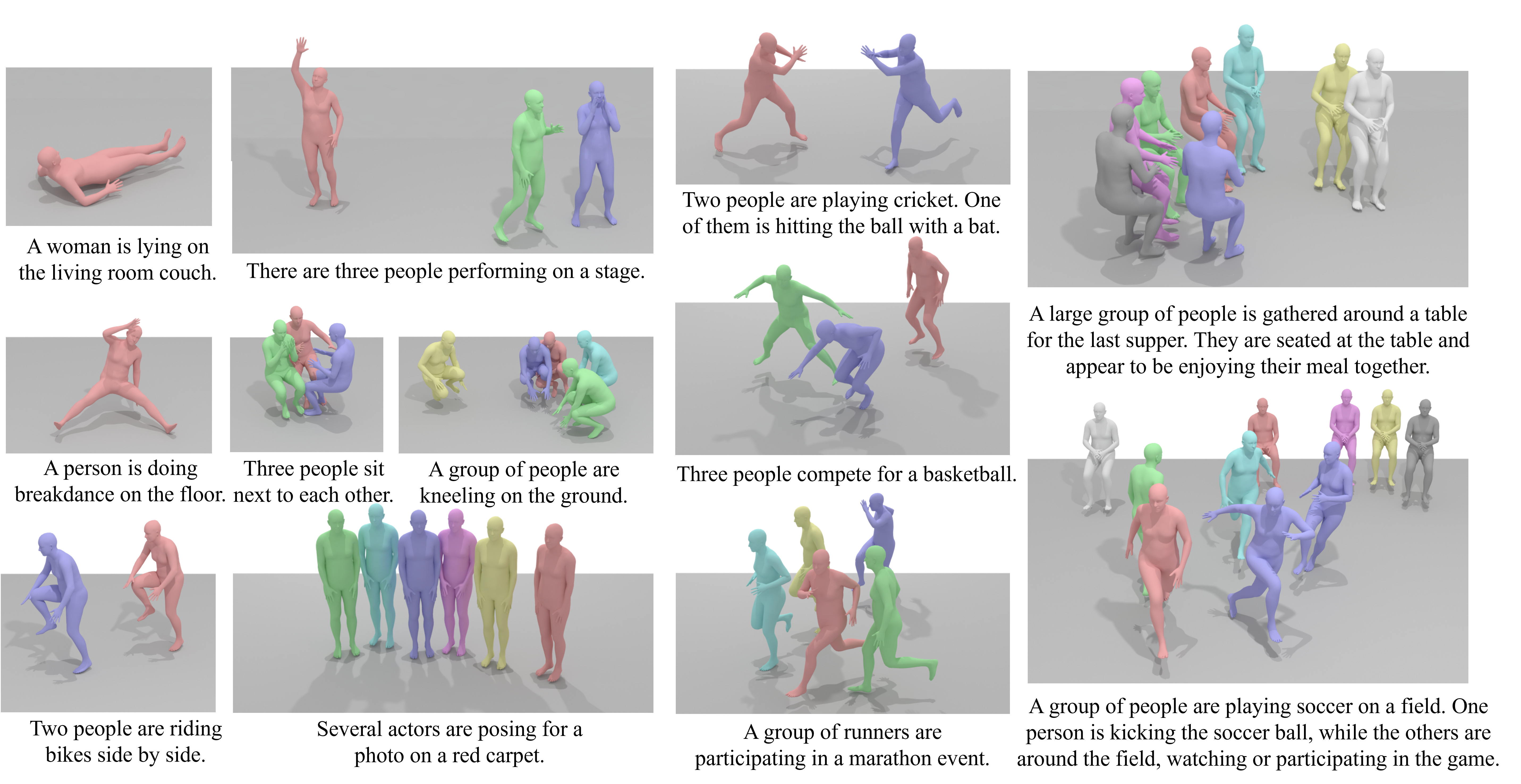}
    \caption{Qualitative result for text-to-pose generation. }
    \label{fig:pose-result}
\end{figure}

\noindent \textbf{Pose Generation.}
Our first stage text-to-pose model can generate diverse, realistic human poses given text prompts shown in \cref{fig:pose-result}. Note that it works well on group prompts out of the distribution of existing motion datasets. Our high quality pose generation module paves the first step for a successful motion generation and pose animation model by providing the fixed middle frame, condition for motion layers, and optionally additional sampling guidance.

\begin{figure}[!ht]
    \centering
    \includegraphics[width=0.95\textwidth]{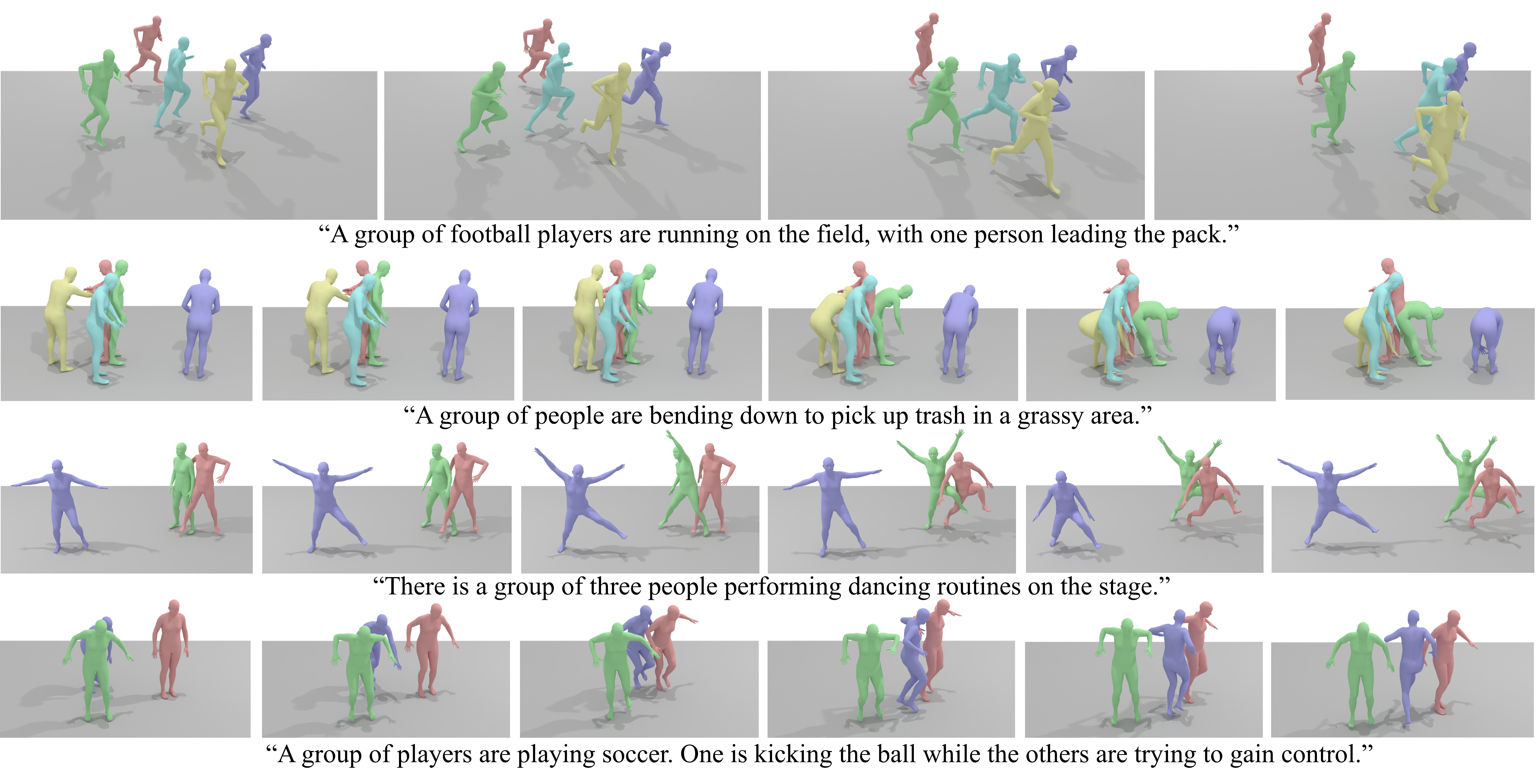}
    \caption{Qualitative results for text-to-motion generation.}
    \label{fig:result}
\end{figure}

\begin{figure}[!ht]
    \centering
    \includegraphics[width=0.95\textwidth]{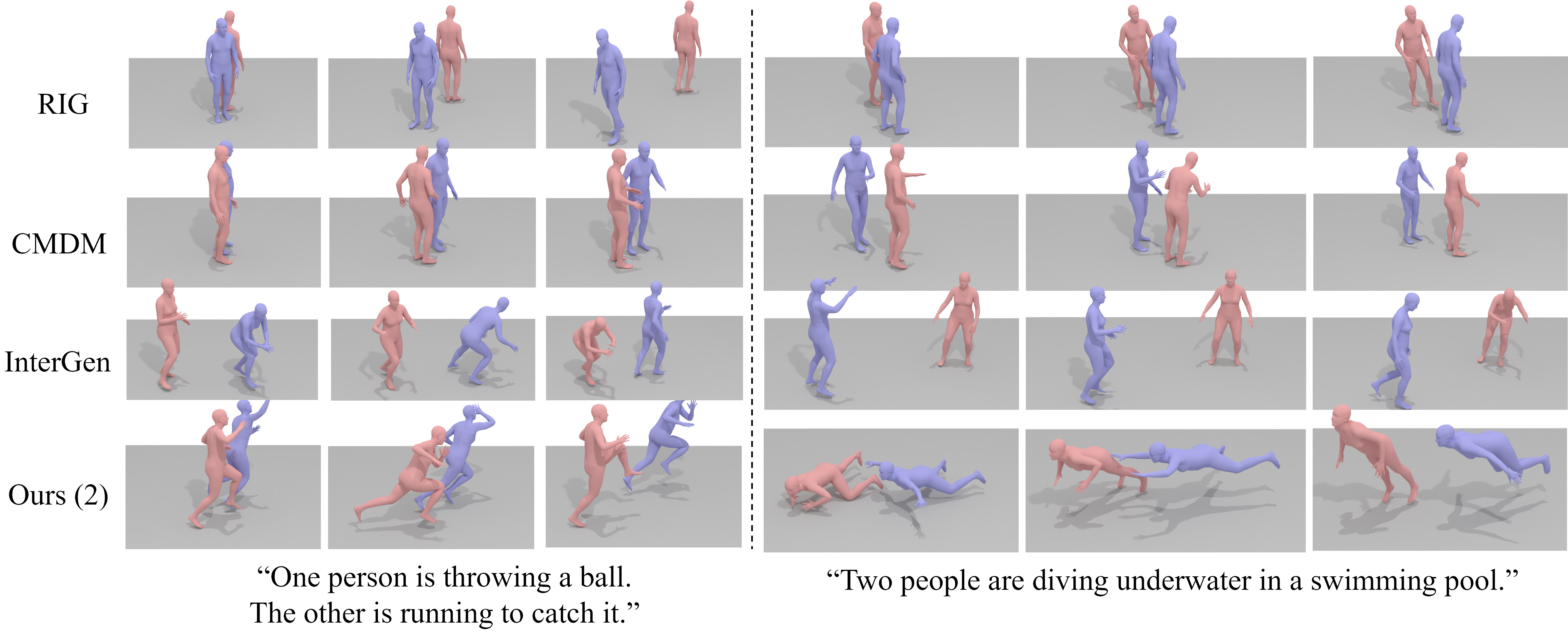}
    \caption{Qualitative comparison with 2-person motion generation baselines.}
    \label{fig:baseline}
\end{figure}

\noindent \textbf{Motion Generation.} In \cref{fig:result}, we visualize a few motion sequences our model generates with the given text prompt. Results show that our model is able to generate high quality motion and interaction for various number of subjects given a wide variety of prompts. In \cref{fig:baseline} we compare our model with baseline results by restricting the sampling subjects to two. For two-person motion, we are able to generate much more realistic motion than RIG~\cite{tanaka2023interaction} and ComMDM~\cite{shafir2023human}, as we take advantage of multi-person pose and motion data in our training instead relying heavily on single-person priors and adding interactive terms. Our model also works with more diverse prompts than InterGen~\cite{liang2023intergen}, benefiting from our joint training strategy that enables training with both high-quality studio captured data as well as in-the-wild motion regression data. 
Readers are encouraged to view the supplementary videos for the animated results. 

\subsection{Quantitative Evaluation}

\cref{tab:main-eval} shows our quantitative results comparing with the naive \textit{Pose-Only} and \textit{Motion-Only} baselines for single-person motion and multi-person pose. Results present the \textit{Pose-Only} and \textit{Motion-Only} results as the upper bounds of performance when solely focusing on per-frame pose quality and per-subject motion quality. Our joint training solution makes a balance between these two extremes, which is able to maintain per frame pose interaction plausibility while adding temporal smoothness throughout the sequences.

\input{tables/main-result}

\cref{tab:baseline} illustrates how our model, when restricted to only two-person motion generation, is able to produce higher-quality 2-person motion results than baselines when conditioned with open-domain text prompts. This is mainly because of the limited distribution of prompts baselines are trained on, and their poor ability to generalize to more in-the-wild textual descriptions that are not constrained by the motion capture studio context.

\input{tables/baseline}

\subsection{Ablation Study}

\noindent We present ablation studies in \cref{tab:ablation} to validate our multiple design choices. 
\input{tables/ablation-pose}

\noindent \textbf{Architecture.} A one-stage model without pre-trained pose results as a condition (A) does not work as well as the two-stage counterpart, because a high-quality pose model can provide strong condition signal to make each frame more plausible. Not freezing the pre-trained pose layers in the joint training stage (B) also does not perform as well, showing that motion data might interfere with the high-quality pose representation learned by pose layers during the first stage. 

\noindent \textbf{Data.} Training without WebVid-Motion data (C) shows a downgrade in text-pose alignments, indicating that multi-person motion data helps improving interaction qualities. Removing all motion texts (D), essentially training the temporal layers unconditionally, also decreases the alignment between pose and text. 


%% file: tables/main-result.tex
 \begin{table*}[h]
    \centering
    \vspace{-10pt}
    \caption{Quantitative metrics comparing our multi-person results with naive baselines. Metrics for real data are evaluated with LAION-Pose and HumanML3D.}

    \scalebox{0.8}{

    \begin{tabular}{l c c c c c c c c c c}
    \toprule
    \multirow{2}{*}{Methods}  & \multicolumn{3}{c}{P-R-Precision $\uparrow$} & \multirow{2}{*}{P-FID $\downarrow$} & \multirow{2}{*}{P-Sim $\uparrow$} & \multirow{2}{*}{P-Div $\rightarrow$} & \multirow{2}{*}{P-MM $\uparrow$} & \multirow{2}{*}{M-FID $\downarrow$} & \multirow{2}{*}{M-Div $\rightarrow$} & \multirow{2}{*}{M-MM $\uparrow$}\\

    \cline{2-4}
    ~ & Top-1 & Top-2 & Top-3 \\
    
    \toprule

        \textbf{Data}    &
         0.621        & 
         0.737        & 
         0.819        & 
         0.000        & 
         0.378        & 
         1.366        & 
         -        & 
         0.000        & 
         1.342        & 
         -         \\

    \midrule

        Pose-Only   & 
         0.678        & 
         0.822        & 
         0.885        & 
         0.077        & 
         0.371        & 
         1.368        & 
         0.903        & 
         0.976        & 
         1.006        & 
         0.667          \\

         Motion-Only   & 
         0.202        & 
         0.307        & 
         0.380        & 
         0.317        & 
         0.157        & 
         1.200        & 
         0.837        & 
         0.613        & 
         1.274        & 
         0.894         \\

    \midrule

        Ours        &
         0.539        & 
         0.704        & 
         0.776        & 
         0.229        & 
         0.304        & 
         1.329        & 
         0.894        & 
         0.684        & 
         1.220        & 
         0.833         \\

    \bottomrule

    \end{tabular}
    }
    \vspace{-15pt}
    \label{tab:main-eval}

\end{table*}

%% file: tables/baseline.tex
 \begin{table*}[h]
    \centering
    \vspace{-5 pt}
    \caption{Quantitative metrics comparing our 2-person results with baselines. The pose and motion metrics for real data are evaluated with LAION-Pose and InterHuman \cite{liang2023intergen}.}

    \scalebox{0.8}{

    \begin{tabular}{l c c c c c c c c c c}
    \toprule
    \multirow{2}{*}{Methods}  & \multicolumn{3}{c}{P-R-Precision $\uparrow$} & \multirow{2}{*}{P-FID $\downarrow$} & \multirow{2}{*}{P-Sim $\uparrow$} & \multirow{2}{*}{P-Div $\rightarrow$} & \multirow{2}{*}{P-MM $\uparrow$} & \multirow{2}{*}{M-FID $\downarrow$} & \multirow{2}{*}{M-Div $\rightarrow$} & \multirow{2}{*}{M-MM $\uparrow$}\\

    \cline{2-4}
    ~ & Top-1 & Top-2 & Top-3 \\
                \toprule

        \textbf{Data (2)}    &
         0.599        & 
         0.722        & 
         0.783        & 
         0.000        & 
         0.378        & 
         1.308        & 
         -        & 
         0.204        & 
         1.218        & 
         -        \\ 

    \midrule

        Pose-only (2)        &
         0.567        & 
         0.754        & 
         0.831        & 
         0.106        & 
         0.382        & 
         1.300        & 
         0.894        & 
         0.953        & 
         1.149        & 
         0.680        \\ 

         Motion-Only (2)   & 
         0.162        & 
         0.260        & 
         0.339        & 
         0.520        & 
         0.191        & 
         1.164        & 
         0.602        & 
         0.558        & 
         1.270        & 
         0.811         \\

         \midrule

        InterGen~\cite{liang2023intergen}   & 
         0.073        & 
         0.126        & 
         0.176        & 
         1.038        & 
         0.113        & 
         0.880        & 
         0.643        & 
         0.734        & 
         1.190        & 
         0.778        \\ 

        RIG~\cite{tanaka2023interaction}   & 
         0.037        & 
         0.072        & 
         0.103        & 
         1.376        & 
         0.061        & 
         0.639        & 
         0.501        & 
         0.925        & 
         1.078        & 
         0.596        \\ 

        ComMDM~\cite{shafir2023human}   & 
         0.043        & 
         0.085        & 
         0.124        & 
         1.160        & 
         0.080        & 
         0.819        & 
         0.714        & 
         0.821        & 
         1.109        & 
         0.765        \\ 

    \midrule

        Ours (2)        &
         0.323        & 
         0.480        & 
         0.591        & 
         0.435        & 
         0.271        & 
         1.329        & 
         0.856        & 
         0.667        & 
         1.213        & 
         0.803         \\

    \bottomrule

    \end{tabular}    }
    \vspace{-10pt}
    \label{tab:baseline}

\end{table*}

%% file: tables/ablation-pose.tex
 \begin{table*}[h]
    \centering
    \vspace{-15pt}
    \caption{Ablation results for different design choices. Bold is the best score. }

    \scalebox{0.8}{

    \begin{tabular}{l c c c c c c c c c c}
    \toprule
    \multirow{2}{*}{Experiments}  
    & \multicolumn{3}{c}{P-R-Precision $\uparrow$} 
    & \multirow{2}{*}{P-FID $\downarrow$} 
    & \multirow{2}{*}{P-Sim $\uparrow$} 
    & \multirow{2}{*}{P-Div $\rightarrow$} 
    & \multirow{2}{*}{P-MM $\uparrow$} 
    & \multirow{2}{*}{M-FID $\downarrow$} 
    & \multirow{2}{*}{M-Div $\rightarrow$} 
    & \multirow{2}{*}{M-MM $\uparrow$}\\

    \cline{2-4}
    ~ & Top-1 & Top-2 & Top-3 \\
    
    \toprule

        \textbf{Data}    &
         0.621        & 
         0.737        & 
         0.819        & 
         0.000        & 
         0.378        & 
         1.366        & 
         -        & 
         0.002        & 
         1.342        & 
         -         \\

    \midrule

    A: One stage        &
         0.386        & 
         0.536        & 
         0.638        & 
         0.407        & 
         0.238        & 
         1.298        & 
         0.802        & 
         0.671        & 
         1.182        & 
         0.745         \\

    B: w/o freeze pose        &
         0.202        & 
         0.307        & 
         0.380        & 
         0.503        & 
         0.157        & 
         1.199        & 
         0.734        & 
         0.613        & 
         1.274        & 
         0.830         \\

    C: w/o WebVid        &
         0.383        & 
         0.534        & 
         0.630        & 
         0.399        & 
         0.246        & 
         1.254        & 
         0.711        & 
         \textbf{0.405}        & 
         \textbf{1.315}        & 
         0.698         \\

    D: w/o motion text        &
         0.313        & 
         0.451        & 
         0.542        & 
         0.368        & 
         0.213        & 
         1.264        & 
         0.749        & 
         0.632        & 
         1.289        & 
         0.804         \\

    \midrule

        Ours        &
         \textbf{0.539}        & 
         \textbf{0.704}        & 
         \textbf{0.776}        & 
         \textbf{0.229}        & 
         \textbf{0.304}        & 
         \textbf{1.329}        & 
         \textbf{0.894}        & 
         0.684        & 
         1.220        & 
         \textbf{0.833}         \\

    \bottomrule

    \end{tabular}   }
    \vspace{-15pt}
    \label{tab:ablation}

\end{table*}

%% file: sections/conclusion.tex
\section{Conclusion}

In this work, we propose the first open-domain text-driven multi-person motion generation algorithm. We design a diffusion-based joint training mechanism with interleaved pose and motions layers, which can utilize multiple data sources simultaneously. We demonstrate diverse and realistic motion generation results  qualitatively and quantitatively superior to baseline methods. We additionally contribute two datasets LAION-Pose and WebVid-Motion, opening more possibilities for future investigation in the field of multi-person motion generation.